\documentclass[]{fairmeta}
\usepackage{wrapfig}

\newcommand{\tautrain}{\tau_{\textup{train}}}

\title{GLoRe: When, Where, and How to Improve LLM Reasoning via Global and Local Refinements}

\author[1,2,*]{Alex Havrilla}
\author[1]{Sharath Chandra Raparthy}
\author[1]{Christoforos Nalmpantis}
\author[1]{Jane Dwivedi-Yu}
\author[3]{Maksym Zhuravinskyi}
\author[1]{Eric Hambro}
\author[1]{Roberta Raileanu}

\affiliation[1]{FAIR at Meta}
\affiliation[2]{Georgia Institute of Technology}
\affiliation[3]{StabilityAI}

\contribution[*]{Work done during Meta internship}

\abstract{State-of-the-art language models can exhibit impressive reasoning refinement capabilities on math, science or coding tasks. However, recent work demonstrates that even the best models struggle to identify \textit{when and where to refine} without access to external feedback. Outcome-based Reward Models (\textbf{ORMs}), trained to predict correctness of the final answer indicating when to refine, offer one convenient solution. However, when used to indicate where to refine, we find that ORMs tend to be \textit{overly-pessimistic} when used to assess intermediate reasoning steps, resulting in excessive refinement of valid solutions. Process Based Reward Models (\textbf{PRMs}), trained to predict correctness of intermediate steps indicating where to refine, have been used to improve LLM reasoning ability via rejection sampling or reinforcement learning (RL) fine-tuning.  But they are expensive to train, requiring extensive human annotations. In this paper, we propose Stepwise ORMs (\textbf{SORMs}) which are trained, only on synthetic data, to approximate the expected future reward of the optimal policy or $V^{\star}$. More specifically, SORMs are trained to predict the correctness of the final answer when sampling the current policy many times (rather than only once as in the case of ORMs). Our experiments show that SORMs can more accurately detect incorrect reasoning steps compared to ORMs, thus improving downstream accuracy when doing refinements. We then train \textit{global} refinement models, which take only the question and a draft solution as input and predict a corrected solution, and \textit{local} refinement models which also take as input a critique indicating the location of the first reasoning error. We generate training data for both models synthetically by reusing data used to train the SORM. We find combining global and local refinements, using the ORM as a reranker, significantly outperforms either one individually, as well as a best of three sample baseline. With this strategy we can improve the accuracy of a LLaMA-2 13B model (already fine-tuned with RL) on GSM8K from 53\% to 65\% when greedily sampled.}

\date{\today}
\correspondence{Alex Havrilla at \email{ahavrilla3@gatech.edu}}

\begin{document}

\maketitle

\section{Introduction}

State-of-the-art large language models (\textbf{LLMs}) exhibit a wide range of downstream capabilities after pre-training. This includes the ability to refine their reasoning on math, science, or coding problems \citep{OpenAI2023GPT4TR, touvron2023llama, Chowdhery2022PaLMSL}. However, under close inspection, this refinement ability is quite brittle, often unable to even identify when a solution needs refinement \citep{Huang2023LargeLM}. When LLMs do produce successful refinements on hard reasoning tasks this is often due to the incorporation of external forms of feedback, e.g. feedback from humans or code, stronger models, or other tools \citep{Zhou2023SolvingCM, Gou2023CRITICLL}. In this work, we carefully examine and improve the self-refinement abilities of LLMs on reasoning tasks without any external feedback other than the ground truth answers of the training problems. Notably, this means we make no use of data or feedback from humans or stronger models. To do so we start by heuristically decomposing the refinement problem into three parts: firstly deciding \textit{when} to refine, then \textit{where} to refine, and finally \textit{how} to refine.


Outcome Based Reward Models (\textbf{ORMs}) \citep{Cobbe2021TrainingVT}, first introduced as an estimator of final answer correctness given a question to do solution reranking, are a natural choice for addressing step one.
For deciding where to refine, we carefully examine the generalization of ORMs to intermediate steps. We find the accuracy of the underlying data generating policy $\pi$ directly affects the ORM's ability to learn correctness of intermediate solutions steps. This leads to the ORM often under-estimating the solvability of a problem from an intermediate step $S_i$. The result is high false-negative rates when used to classify steps with errors. Process Based Reward Models (\textbf{PRMs}) instead are trained to directly estimate the correctness of each step. Yet this requires extensive human labeling of model-generated solution steps as valid or invalid. In an effort to improve our ability to give intermediate step feedback, we introduce the Stepwise ORMs (\textbf{SORMs}) which explicitly predict labels at each step indicating the presence of an error. We generate SORM training data by sampling a student policy $\pi$ many times at a step $S_i$ in solution $S$, labeling $S_i$ as valid if we successfully reach the final answer. From an RL perspective, this can be interpreted as learning (a lower bound of) the optimal value function $V^*$ of the reasoning task via approximation of the optimal policy $\pi^*$ with rejection sampling. The resulting SORM gives better intermediate step-level feedback, allowing us to give information to the refinement model about both \textit{when} and \textit{where} to refine. The refinement model must then only decide \textit{how} to refine.

We initially train \textit{global} refinement models capable of refining the entire reasoning trace without any feedback beyond an initial draft solution $D$. The training data is generated synthetically, by pairing correct solutions with incorrect solutions as in \citet{Welleck2022GeneratingSB}. An evaluation of the global refinement model confirms its inability to correctly identify when to refine, demonstrating the need for an ORM. Reusing the SORM training data, we train a \textit{local} refinement model which uses the feedback given by the SORM to identify the first incorrect reasoning step. 
We then compare the performance of global versus local refinements on a test set of incorrect solution drafts, finding similar refinement accuracy but on largely disjoint sets of problems. In this sense the global and local refinement models are complementary, with local refinements often able to solve problems global refinements cannot and vice versa. To obtain our best results we combine both global and  local refinements, using the ORM to choose the most promising one by acting as a reranker of both plus the initial draft. Using this strategy, we can improve the accuracy of an already strong RL fine-tuned Llama-2 13B mode from 53\% to 65\% when greedily sampled.


In summary we make the following contributions:

\begin{itemize}
    \item Decompose the refinement problem into three parts, namely deciding \textit{when, where, and how} to refine a solution by leveraging reward models (RMs).
    \item Highlight the limitations of ORMs in judging the correctness of intermediate steps, despite their ability to judge the correctness of the final answer.
    \item Introduce the step-wise ORM (SORM) to refine which is trained only on synthetic data and can more accurately evaluate intermediate steps than the ORM. 
    \item Propose a new method for refining LLM reasoning that decides \textit{when} to refine using an ORM, \textit{where} to refine using a SORM, and \textit{how} to refine using both global and local refinements. We find the two types of refinement are complementary, each able to solve a large class of problems the other cannot.
    \item Demonstrate performance improvements of up to 12\% on GSM8K for a 13B LLaMA-2 model using our approach.
\end{itemize}


\section{Background}
\label{sec:background}

\textbf{Reasoning:} We define a reasoning task $\tau$ as a distribution of (natural language) question/answer pairs $(Q, A) \sim \tau$. The answer could be either a single final answer, typically a numerical value in case of math problems for ease of evaluation, or include a CoT style solution trace justifying a numerical final answer. We often further write the answer $A$ as consisting of atomic steps $A = (S_1, ..., S_L)$ with the final answer being given on step $L$. The notion of a start of a new "step" is problem dependent but in our case always corresponds to a newline token.



\textbf{Reward Modeling:} Given a reinforcement learning (RL) environment, a reward model can be trained to approximate the reward coming from an action $a$ in state $s$ \citep{christiano2017deep}. In the language setting, reward models are trained to approximate the reward given to a response generated by a LLM \citep{Ouyang2022TrainingLM}. The reward is generally sparse and given at the end of a generation as in the case of RLHF \citep{christiano2017deep, ziegler2019fine} where a contrastive preference model is learned for RL and rejection sampling. 

Similar to this is the \textit{Outcome-based Reward Model} (\textbf{ORM}) first proposed as a final answer verifier used to rerank GSM8K solutions \citep{Cobbe2021TrainingVT}. Formally, we say the ORM estimates $p(\texttt{is\_correct}(A) | Q, A)$ where $Q$ is a question and $A$ is a model generated answer. Training data for the ORM is generated by sampling an underlying student model $\pi$ many times on questions from a reasoning task $\tau$. The ORM is then trained to predict $p(\texttt{is\_correct}(A) | Q, P_i)$ where $P_i$ is prefix of intermediate steps $(S_1, ..., S_i)$ and $A$ is any hypothetical continuation of $P_i$ sampled from $\pi$. i.e., at intermediate steps we may interpret the ORM as estimating the probability of $P_i$ leading to the correct final answer. We may sometimes write $ORM_\pi$ to emphasize the ORM's dependence on its data generating student model $\pi$. More recently, \textit{Process-based Reward Models} (\textbf{PRMs}) have been proposed to directly supervise the correctness of each step in a solution $A = (S_1,...,S_L)$ \citep{Lightman2023LetsVS, Uesato2022SolvingMW}. Formally, we write a PRM predicts $p(\texttt{is\_correct}(S_i) | P_i, Q)$ where $S_i$ is the last step of $P_i$.

\textbf{Refinement:}
We define a refinement of a draft solution $A_D$ and question $Q$ as a new solution $A_R$ generated by conditioning on both $Q$ and $A_D$. We consider both global refinement models, which take as input only $Q, A_D$ and predict $p(A_R | Q, A_D)$, and local refinement models, which take as input an extra parameter $E$ indicating the location of an error in $A_D$, to predict $p(A_R | Q, A_D, E)$.

\textbf{Notation:} 
For the rest of the paper we refer to the pre-trained LLM fine-tuned for downstream tasks as the \textit{base model}. We fine-tune the base model, either on supervised data or using RL, to produce a student model that generates answers $A$ given a question $Q$. Sometimes we may also write the student model as a policy $\pi$ implicitly depending on learnable parameters $\theta$. $\mathcal{D}_\textup{TASK}$ will be used to denote a dataset for TASK $\tau$ with train split $\mathcal{D}_\textup{TASK}^\textup{train}$ and test split $\mathcal{D}_\textup{TASK}^{\textup{test}}$ being implicit. We will use $Q$ to denote a question and $A_1,...,A_k$ to denote solution traces. Sometimes we will write $A = (S_1,...,S_L)$ which decomposes the solution trace $A$ into intermediate steps $S_i$. $P_i = (S_1, ..., S_i)$ will be used to denote the prefix of steps up to $S_i$. Additionally we will sometimes use $A_{GR}$ and $A_{LR}$ to represent global and local refinements of $A_D$. $V^\pi$ denotes the value function of policy $\pi$. $V^*$ denotes the optimal value function with dependence on the background task implicit.


\section{Related Works}
\textbf{LLM Reasoning:} State-of-the-art (SOTA) large language models (\textbf{LLMs}) \citep{OpenAI2023GPT4TR, touvron2023llama, Bai2022ConstitutionalAH, Chowdhery2022PaLMSL} demonstrate increasingly impressive abilities on hard reasoning tasks as studied by a wide range of math, science, and code benchmarks \citep{Cobbe2021TrainingVT, Hendrycks2021MeasuringMP, Sawada2023ARBAR, Liang2022HolisticEO, Srivastava2022BeyondTI, Rein2023GPQAAG, Mialon2023GAIAAB, Chollet2019OnTM, hendrycks2021measuring, Austin2021ProgramSW, Mishra2022LilaAU, patel2021nlp, eval-harness}. \textit{Chain of thought} (\textbf{CoT}) \citep{Wei2022ChainOT} and related techniques \citep{Chen2022ProgramOT, Yao2023TreeOT, Besta2023GraphOT} have emerged as dominant methods significantly boosting LLM performance on these types of tasks. CoT methods allow LLMs to defer giving their final answer by first generating a "chain of thought" involving intermediate computations needed to correctly solve the problem.

\textbf{LLM Refinement:} Intimately related to reasoning ability is a model's ability to refine previous answers. This work studies the ability of large language models to self-refine their CoT solutions to math reasoning tasks. Several works \citep{Yao2022ReActSR, madaan2023selfrefine, Zhou2023SolvingCM} demonstrate SOTA LLM self-refining and self-critiquing abilities on a range of tasks via prompting and/or tool usage. However, recent work \citep{Huang2023LargeLM} argues even for the strongest models such techniques struggle on hard, open-ended reasoning tasks where the model itself must decide when to stop refinement.


Other papers use hand-crafted data augmentation \citep{Paul2023REFINERRF} or gather human data \citep{wang2023shepherd, Chen2023ImprovingCG, Lee2023PlatypusQC, Saunders2022SelfcritiquingMF, Schick2022PEERAC} while still others use techniques from reinforcement learning to generate critiques \citep{Akyurek2023RL4FGN, Yao2023RetroformerRL} for larger models. Most related to us is \citep{Welleck2022GeneratingSB} which trains global refinement models in an implicit reinforcement learning like manner by pairing \textit{low-value} rollouts with \textit{high-value} rollouts.

Process-based reward modeling (\textbf{PRMs}) \citep{Uesato2022SolvingMW, Lightman2023LetsVS} gives a denser, step-by-step reward for the "correctness" of a particular step without explicitly modeling the step's impact on the correctness of the final answer. Both ORMs and PRMs are most often used as rerankers over large numbers of candidate solutions, with PRMs generally outperforming ORMs \citep{Lightman2023LetsVS}. However, PRMs areexpensive to train, requiring extensive human annotation of each step. \citet{Uesato2022SolvingMW} directly compares the performance of a 70B ORM vs PRM on GSM8K, finding both performing similarly when used as a reward for RL and for reranking. They qualitatively note the ORM appears to somewhat generalize to intermediate steps in a manner similar to a PRM but do not quantitatively ablate this observation over multiple models or tasks. \citet{Li2022MakingLM} attempt to train synthetic stepwise verifiers similar to a PRM which are then used for Monte Carlo Tree Search. Concurrent work \citep{Wang2023MathShepherdVA} proposes training a synthetic process based reward model in a manner similar to our SORM. They then use the RM downstream for RL fine-tuning and rejection sampling.

In contrast to the above works we conduct a careful comparison of ORM/SORM verification abilities at the step level. We then propose to utilize the ORM/SORM for refinement. We accomplish this by generating fully synthetic stepwise labels which allow us to train both the SORM and refinement models.


\section{Method}
\label{sec:method}

\begin{figure}
    \centering
    \includegraphics[width=\columnwidth]{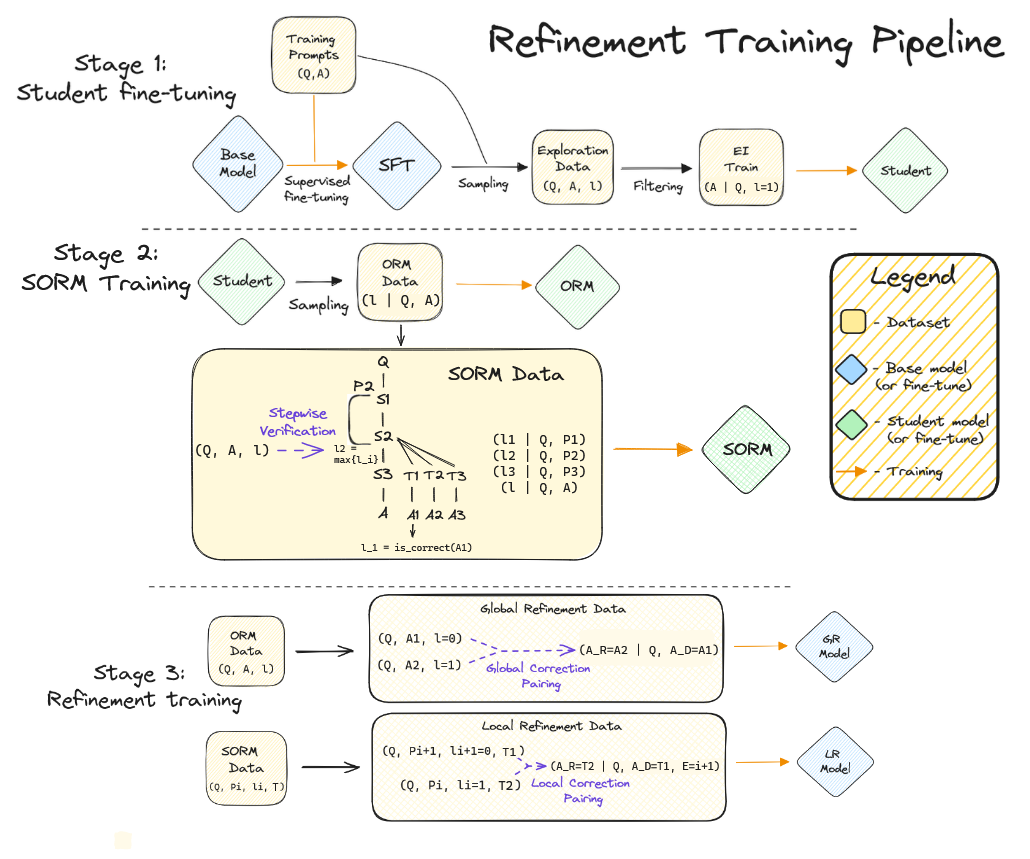}
    \caption{Diagram for three-stage refinement training pipeline. First we RL fine-tune the base model to produce a strong student policy $\pi$. Then we generate ORM/SORM training data by sampling $\pi$ on the training data. Finally, we generate refinement data by pairing together incorrect rollouts with correct rollouts globally and locally. Note, $(Q, A, l)$ denotes a question, answer pair with binary correctness label $l$. A SORM training sample $(Q, P_i, l_i, T)$ includes a prefix of steps $(S_1,...,S_i)$, a binary correctness label $l_i$ for the prefix, and the set of verification rolluts $T_1,...,T_K$ from $P_i$ verifying correctness of $P_i$. Global correction pairing is used to produce global refinement training data by pairing incorrect ORM rollouts with correct ORM rollouts. Analgously, local correction pairing pairs incorrect verifictions $T_-$ of (incorrect) $P_{i+1}$ with correct verifications $T_+$ of $P_i$. This then produces a label $E = i+1$ indicating an error at step $i+1$ in the initial draft $A_D=T_-$ and refinement $A_R = T_+$.}
    \label{fig:refinement-train-pipeline}
\end{figure}

We start by decomposing the refinement problem into three stages: First, learning \textit{when} a draft $D$ is correct and when it needs refinement. Second, learning \textit{where} to begin refinement by identifying the first incorrect step. Third, learning \textit{how} to correct the initial draft. We can naturally address step one by using the ORM which is trained to predict the probability of a draft being correct. This alleviates some of the difficulty, now only requiring the refiner to identify where and when to refine. Additionally, when doing local refinement, we propose using the (S)ORM to localize the position of the first error. This simplifies the task even more, as now the local refiner must only decide how to fix the error and continue from there.

\textbf{Localizing errors with Reward Models:} To identify errors at the step level we can leverage the ORM by taking its intermediate prediction $ORM_\pi(Q, P_i)$ at a step $S_i$ where $P_i = (S_1, ..., S_i)$ is the prefix of all steps up to $S_i$. Recall the ORM is trained to predict the likelihood a solution with prefix $P_i$ results in a correct final answer. Importantly, the likelihood inferred from this training data is heavily dependent on the data generating policy $\pi$. For this reason we sometimes include the subscript $ORM_\pi$, omitting it when not needed.

To best understand the behavior of the ORM's prediction at an intermediate step $S_i$, we can interpret it as the \textit{value function} of $\pi$. Recall the value function $V^\pi$ of a policy $\pi$ is computed as $V^\pi(S) = \mathbb{E}_{\tau \sim \pi(S)} R(\tau)$ i.e. the mean return of the policy $\pi$ from the state $S$. In the context of reasoning problems, the states we consider are of the form $S = (Q, S_1,..., S_i)$ with question $Q$ and intermediate steps $S_j$. In our setting by default there is only a sparse reward of $+1$ given at the terminal state for a correct final answer.

We can write $ORM_\pi(Q, P_i) \approx p(\texttt{is\_correct(\textup{A})} | Q, P_i, \pi)$ where $P_i = (S_1, ..., S_i)$ is the prefix of all prior steps and $\texttt{is\_correct(A)}$ is the event that a full solution $A$ sampled from $\pi$ with prefix $P_i$ has the correct final answer. We can then write $\mathbb{E}_{A \sim \pi(Q, P_i)} R(A) = \mathbb{E}_{A \sim \pi(Q, P_i)} 1_{\texttt{is\_correct(A)}} = p(\texttt{is\_correct(A)} | Q, P_i, \pi)$. Therefore, an approximation to the value function of a policy $\pi$ is predicting exactly the same thing as the outcome-based reward model at an intermediate step $S$. So we may treat the \textbf{ORM as approximating a value function for the student model} $\pi$ used to generate its training data.


Ideally we might want to use the ORM to identify where a mistake was made by finding the first step $S_i$ such that $ORM(Q, P_i) \leq 0.5$ i.e. $P_i$ is likely to result in the wrong answer. However, because the ORM is acting as a value function for $\pi$, it tends to \textbf{hallucinate error steps} simply because it expects the data generating student $\pi$ to fail. For example, if $\pi$ almost always fails problems involving division, the ORM will assign low probability of success to a division problem even before the student takes its first step. In these cases we say the ORM is \textit{overly pessimistic}. This is not ideal when using the ORM to identify the location of mistakes. 




\textbf{Learning a Step-Wise ORM (SORM):} Another natural candidate which could be used to identify mistakes at each step is a Process Based Reward Model (PRM) \citep{Lightman2023LetsVS}. A PRM estimates the probability of correctness of a step $S_i$, $p(S_i\texttt{ correct}|Q,S_1,S_2,...,S_i)$ independently of its impact on the final answer. However, this would be expensive, requiring collecting human annotated samples. Instead, we propose to approximate the \textit{optimal value function} $V^*$ of the reasoning task. $V^*$ corresponds to the value function of the \textit{optimal policy} which  is able to successfully solve the reasoning task from any logically valid intermediate state $S_j$. Such an optimal value function would have $V^*(Q, S_1,..., S_i) = 1$ for a solution prefix with no mistakes, and $V^*(Q, S_1, ..., S_i) = 0$ if the prefix already contains a mistake which will result in an incorrect final answer. 
We call models we train to directly approximate $V^*$ stepwise ORMs or \textbf{SORMs}.

As discussed in \citet{Uesato2022SolvingMW}, the ORM possesses some knowledge of intermediate solution correctness, allowing it to approximate a PRM. However, we find in practice this property is dependent on the size of the base model and the difficulty of the task $\tau$, with ORMs trained on data from larger students and easier tasks giving better approximations to a PRM. When interpreting the ORM as a value function $V^\pi$ of the data generating student, this makes sense. A larger, more capable student will better approximate the optimal policy $\pi^*$, resulting in a better approximation of the ORM to $V^*$.


\subsection{Training pipeline}

Recall, we assume no access to data from humans or better models for fine-tuning. Thus we must generate all training data synthetically for both global and local refinement. Additionally we must generate data for both the ORM and SORM. We divide our proposed training pipeline in three steps. See Figure \ref{fig:refinement-train-pipeline} for a diagram outlining each step.

\textbf{Step 1: Fine-tuning a student model}

To produce base checkpoints from which we can generate ORM/SORM training data and initial refinement drafts $A_D$ we fine-tune models using Expert Iteration (\textbf{EI}) \citep{Silver2017MasteringCA}. This is done by sampling the student model $K = 96$ times per question and filtering out rollouts with incorrect final answers. De-duplication is then performed on the remaining samples to construct a new finetuning dataset $\mathcal{R}_1$. We then combine this with any available SFT data producing $\mathcal{D}_1$ which we use to again fine-tune the pre-trained model. This process is repeated until the maj@1 score of each subsequent fine-tune converges. Note, the fine-tuning dataset used at step $i$ is $\mathcal{D}_i = R_i \cup \mathcal{D}_{i-1}$: the union of rollouts generated at the $ith$ step with previously generated training data ($\mathcal{D}_0 = \emptyset \textup{ or } SFT$). In the case of GSM8K we first fine-tune each pre-trained model on the given supervised fine-tuning (\textbf{SFT}) data. For SVAMP, which has no CoT SFT data, we 1-shot prompted the pretrained model to generate solutions used to construct an initial EI dataset. We call the resulting model the student model or student policy $\pi$. For more details of this training process and resulting models see Section \ref{sec:rl_reasoning} in the appendix.

\textbf{Step 2: Training the ORM/SORM}

We generate ORM training data by sampling the RL fine-tuned student policy $\pi$ $K$ times per prompt. As usual, we then label each intermediate step $S_i$ as correct if the final answer is correct and incorrect otherwise. To generate training data for our SORM we sample an approximation of the optimal policy $\pi^*$ at each step $S_i$ in a model generated solution and check correctness of the final answer. We aim to approximate $\pi^*$ via rejection sampling of our student policy $\pi^*$. Concretely, to produce a training label for a step $S_i $ in model generated rollout $S$, we sample the student policy $\pi$ for $K$ rollouts starting from the prefix $P_i = (S_1, ..., S_i)$. This produces verifying traces $T_1,..., T_K$ with correct final answers indicated by $l_1,..., l_K$. We then label $S_i$ as $\texttt{positive}$ if $\max_j l_j = 1$ i.e. we can find the correct final answer starting from $S_i$. In practice we sample $K = 8$ rollouts per step, each generating at most 300 tokens. Otherwise we label $S_i$ as $\texttt{negative}$. We then train the SORM in exactly the same manner as the ORM, predicting the appropriate label after each step in a solution. See Section \ref{sec:sorm_data_acc} for a comparison of the labels assigned by this process to ground truth human labels.



\textbf{SORM data post-processing} To improve our approximation to the optimal policy via rejection sampling we apply several post-processing steps: \textbf{1)} If a step $S_i$ has a positive label $l_i$ we set $l_j = 1$ for $j \leq i$. I.e. all steps before a positive steps are labeled as positive. This accounts for particularly hard problems where the student is able to find the solution with $K$ samples from the step $S_i$ but not any prior step $S_j$, $j < i$. \textbf{2)} We enforce a \textit{consistency constraint} on the verifying rollouts, requiring each intermediate result $R_i$ computed on step $S_i$ of the solution to be used later on. This helps prevent false positives by requiring a verification to make full use of the previous steps it's verifying. In practice we implement this by checking for each $R_i$ as a string in the suffix after $P_i$. \textbf{3)} We balance the number of positive and negative labels at each prefix length in the training dataset. This is crucial, as otherwise there is an imbalance of positive labels towards the start of solutions and negative labels towards the end. This imbalance is easy for SORMs to exploit, leading to models which almost always predict a $\texttt{positive}$ label in the first few steps a $\texttt{negative}$ label towards the end.

As an additional baseline we consider the \textbf{Balanced ORM} which simply balances the number of positives and negatives per question in the ORM training dataset. This is done in an attempt to mitigate the overly pessimisstic behavior of the ORM described earlier.

Our SORM approximation is motivated by observations from concurrent work 
which shows our student $\pi$ does not need to engage in too much exploration, i.e. sampling, to solve most problems sufficiently in distribution of pretraining data. This suggests rejection sampling to be capable of providing a decent approximation to the optimal policy. Additionally, the deterministic dynamics of the reasoning environment allows us to only sample once from the optimal policy $\pi^*$ to compute $V^*$ at a prefix $P_i$. This further reduces our sampling requirements, while also allowing us to conclude that if rejection sampling can solve the problem from a prefix $P_i$, then $\pi^*$ will also solve the problem from $P_i$. Note of course rejection sampling will be weaker than $\pi^*$, resulting in the SORM being an under-approximation of $V^*$.


\begin{figure}
    \hspace{-1.1cm}
    \includegraphics[scale=0.27]{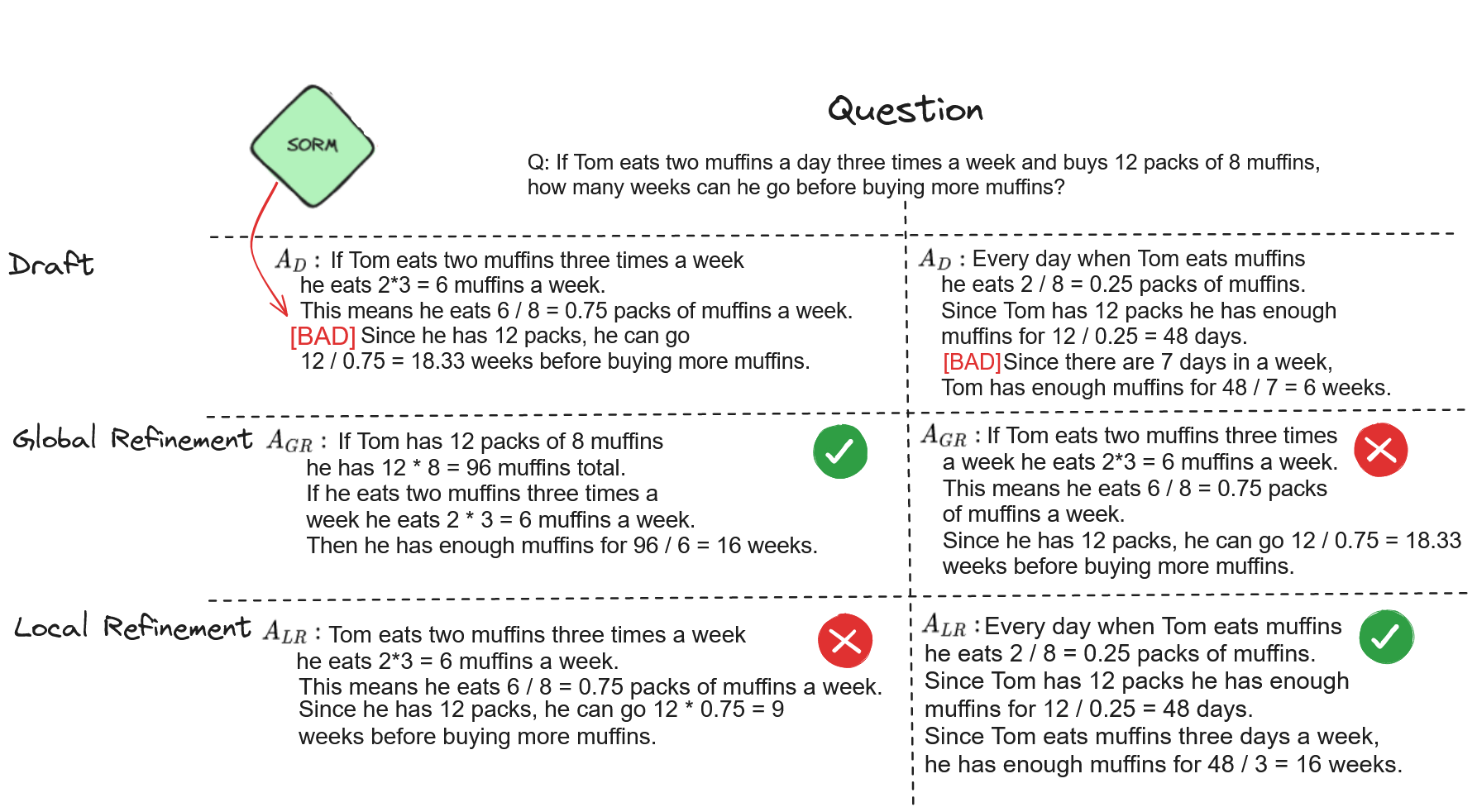}
    \caption{Example of local and global refinements on a math word problem. \textbf{Left:} The local refinement does poorly with a student which struggles dividing by a fraction. Although all prior steps leading up to the fractional division are valid, the local refinement model is forced to either attempt the difficult operation again or choose the wrong operation entirely. In contrast, the global refinement model may attempt to solve the problem with an entirely new approach. \textbf{Right:} In this draft, the model is very close to the final answer, only making a simple mistake at the end. The local refinement is able to correct this simple mistake. In contrast, the global refinement must start from scratch.}
    \label{fig:refinements}
\end{figure}

\textbf{Step 3: Training refinement models} 

To train a local refinement model we need a dataset of the form $(Q, A_D, A_R, E)$ where $Q$ is a question, $A_D$ is an initial draft, $E$ labels the location of the first error in $A_D$ indicating where to refine, and $A_R$ is a refinement with the correct final answer. In pratice, $E$ is communicated to the local refinement as a ``[BAD]'' token prefixing the incorrect step $S_i$ in the draft. Then, at test time, we need a model predicting $p(E|Q,A_D)$ to localize errors in the draft. Conveniently, we explicitly train the SORM to predict the correctness of each step in $A_D$. Thus, to produce $E$ we infer the SORM on all steps and return the index of the first step with predicted correctness below a threshold $T$. Further, \textbf{we can construct a refinement training dataset with error annotations using the SORM dataset}. Given an incorrect model rollout $A = (S_1, S_2, ..., S_L)$ we can locate step $S_i$ as containing the first error by identifying $l_i = 0$ as the first zero label in the trace. We then pair $A$ with a correct verifying trace $T$ from the previous (correct) step $S_{i-1}$. This creates a training pair $(A, T)$ where we label the first error in $A$ as $E = i$. See Figure \ref{fig:refinements} for an example.

We construct a dataset for global refinement similarly using the ORM training dataset. This is done by pairing incorrect rollouts $A_{\textup{incorrect}}$ with correct rollouts $A_{\textup{correct}}$ for the same question $Q$. This constructs a training tuple $(Q, A_{\textup{incorrect}}, A_{\textup{correct}})$. To maintain a format similar to local refinement, we put a $[BAD]$ token at the very start of the incorrect rollout. We combine both refinement datasets to train a model capable of both global and local refinement.

\subsection{Evaluation}

We construct a test set for both the ORM/SORM and refinement models by sampling the student model greedily on test questions $Q$ from the task $\tau$. For each benchmark this gives us a test set with prompts of the form $(Q, A_D)$ where $Q$ is the problem and $A_D$ is an initial draft. For both benchmarks we refer to this as the $(Q, D)$ test set. To generate intermediate step labels we use the same process as used to generate SORM training data. We evalaute the ORM and SORM on this test set by comparing their predictions to these ground truth labels. 

To evaluate the global refinement performance we greedily infer the refiner on each $(Q, A_D)$ sample and compare the resulting refinement $A_{\textup{GR}}$ to the ground truth. To evaluate the local refinement model we first annotate each $(Q, A_D)$ pair with the location of its first error using the ORM or SORM. This forms a $(Q, A_D, E)$ triplet which we use to greedily sample the local refiner.

\begin{figure}
    \centering
    \includegraphics[scale=0.4]{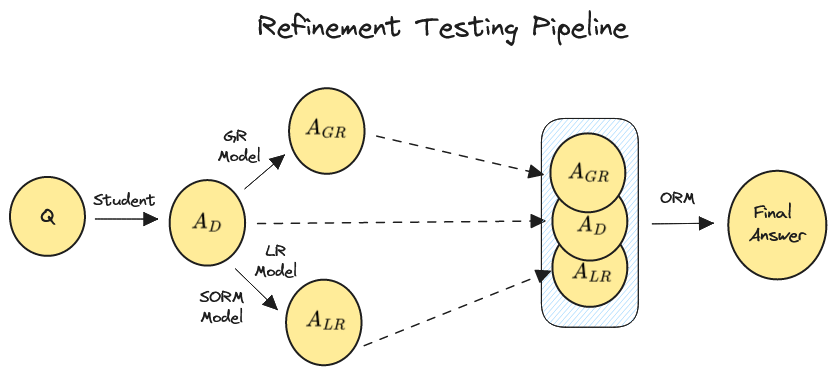}
    \caption{\textbf{Evaluation Pipeline} for global and local refinement models. We first sample a draft $A_D$ from the student model then sample global and local refinements. The ORM is then used to determine which response to select as the final answer among these three candidate solutions.}
    \label{fig:refinement-test-pipeline}
\end{figure}

For our best results, we propose to sample both a global refinement $A_{\textup{GR}}$ and a local refinement $A_{\textup{LR}}$ for a draft $A_D$ and choose the best solution using the ORM reranker. This strategy stems from our observation that global and local refinements each solve complementary, partially non-overlapping subsets of problems the student initially fails on. Thus combining both refinements with the draft significantly expands the set of problems we can solve. Additionally, using the ORM to rerank refinements allows for a cleaner comparison against a best-of-three baseline from the draft-generating student $\pi$. See Figure \ref{fig:refinement-test-pipeline} for a diagram of the evaluation pipeline.


We also highlight more exploratory work in the appendix. In the main body we consider only \textit{process-based} local refinement, which relies on locating reasoning errors in a solution trace. One drawback of this approach is its agnosticism to the abilities of the student model doing refinement. Alternatively, we consider \textit{value-based} refinement which relies on feedback identifying the step in a solution from which the model has the best chance of succeeding. A comparison to process-based refinement is done in appendix Section \ref{sec:value_refinement}. Additionally, in appendix Section \ref{sec:rl_global_refinement}, we compare refinement training using expert iteration to other RL algorithms with various reward schemes.

\section{Results}

We evaluate our refinement pipeline on the GSM8K \citep{Cobbe2021TrainingVT} and SVAMP \citep{patel2021nlp} math word problem benchmarks. We fine-tune Llama-2 7B and 13B to produce all downstream models including the ORM, SORM, and refinement models. Note, the evaluation of each model size is self-contained, not utilizing any data or feedback from models of a different size. maj@1 model scores via greedy sampling will be used to evaluate model performance. Hyperparamters for each phase of training are supplied in Section \ref{sec:hparams} of the appendix.


\begin{table}[h]
\begin{minipage}{0.45\textwidth}
 \centering
    \begin{tabular}{@{}lrrrr@{}}
    \toprule \\
    & \multicolumn{2}{c}{GSM8K} & \multicolumn{2}{c}{SVAMP} \\
    \cmidrule(lr){2-5}
    & 7B & 13B & 7B & 13B \\
    \midrule
    ORM & 0.74 & 0.73 & 0.77 & 0.85 \\
    Balanced ORM & 0.73 & 0.74 & 0.77 & 0.83 \\
    SORM & \textbf{0.79} & \textbf{0.81} & \textbf{0.78} & \textbf{0.87} \\
    \end{tabular}
    \caption{Step-level accuracy of 7B/13B ORM and SORM on test set labels. Note: the test sets are well balanced with positive labels representing 45\%-55\% of samples. The SORM has better step level accuracy than ORM on the harder GSM8K benchmark but comparable step level accuracy on SVAMP.}
    \label{tab:dense_label_acc}
\end{minipage}
\hfill
\begin{minipage}{0.45\textwidth}
 \centering
    \begin{tabular}{@{}lrrrr@{}}
    \toprule \\
    & \multicolumn{2}{c}{GSM8K} & \multicolumn{2}{c}{SVAMP} \\
    \cmidrule(lr){2-5}
    & 7B & 13B & 7B & 13B \\
    \midrule
    ORM & \textbf{0.82} & \textbf{0.85} & \textbf{0.75} & \textbf{0.82} \\
    Balanced ORM & 0.8 & 0.82 & 0.73 & 0.79 \\
    SORM & 0.79 & 0.8 & 0.74 & 0.79 \\
    \end{tabular}
    \caption{Final answer accuracy of 7B/13B ORM and SORM on test set labels. Note: the test sets are well balanced with positive labels representing 45\%-55\% of samples. The ORM has better accuracy than the SORM at predicting final answer correctness.}
    \label{tab:final_label_acc}
\end{minipage}
\end{table}


\subsection{Evaluting the ORM and SORM}

\textbf{SORMs are better than ORMs at evaluating intermediate answers:} On GSM8K the SORM improves over the intermediate step accuracy of the ORM by up to 8\% from 73\% to 81\% (See Table \ref{tab:dense_label_acc}). This confirms the ORM does a reasonable job estimating intermediate step correctness but can still be improved, particularly for smaller models on a hard tasks like GSM8K. We'll see this difference in label accuracy also translates into a difference in refinement final accuracy, where it is critical for the ORM/SORM to reliably identify locations of mistakes. In comparison, the balanced ORM underperforms, having comparable intermediate accuracy to the ORM. This is despite qualitiatively appearing to fix the ORM's over-pessimism, as the balanced ORM assigns roughly 50\% chance of success to all questions. We also examine the types of errors models make, finding the SORMs to have a balanced numbers of false positives and negatives when using a 0.5 as the classification threshold.

\textbf{ORMs better approximate $V^*$ on easier tasks:} On SVAMP the ORM has better step accuracy than on GSM8K (see Table \ref{tab:dense_label_acc}), particularly the 13B model. As a result the SORM offers less improvement. Most questions in GSM8K are relatively more difficult, requiring at least 4 steps to solve. In contrast, most questions in SVAMP require at most three key steps. This small number of steps likely makes it easier for the ORM to generalize. Additionally, the EI models trained on SVAMP reach on average 15\% higher accuracy than the same sized model on GSM8K. This makes the base student model a closer approximation to $\pi^*$ on SVAMP, making the ORM a closer approximation to $V^*$. 

The importance of a strong data generating student $\pi$ is further highlighted by the difference in accuracies between 7B and 13B models on SVAMP. The 7B student EI model gets an accuracy of 58\%, whereas the 13B model gets an accuracy of 70\%. Correspondingly, the 13B ORM model performs much better at on intermediate steps than the 7B model. Yet in contrast the 13B ORM on GSM8K performs slightly worse at intermediate steps than 7B. This is perhaps partially explained by the performance of the 13B EI student on GSM8K which only improves 5\% over the 7B student.

\textbf{ORMs are better than SORMs at evaluating final answers:} Despite the SORM being generally better at predicting intermediate steps, it is slightly worse at predicting final answer correctness compared to the ORM. This is true for both benchmarks, with the 13B SORM on GSM8K lagging by 5\% (See Table \ref{tab:final_label_acc}). However, part of this difference is likely due to statistical biases the ORM is able to exploit, improving final answer accuracy at the cost of over-pessimism. For example, if the problem involves division, the ORM knows the student is likely to fail and immediately predicts a low probability of success. In contrast the SORM is forced to be more optimistic, attempting to carefully examine the correctness of each intermediate step.

\begin{wrapfigure}{r}{0.6\textwidth}
\centering                  
    \includegraphics[width=0.6\textwidth]{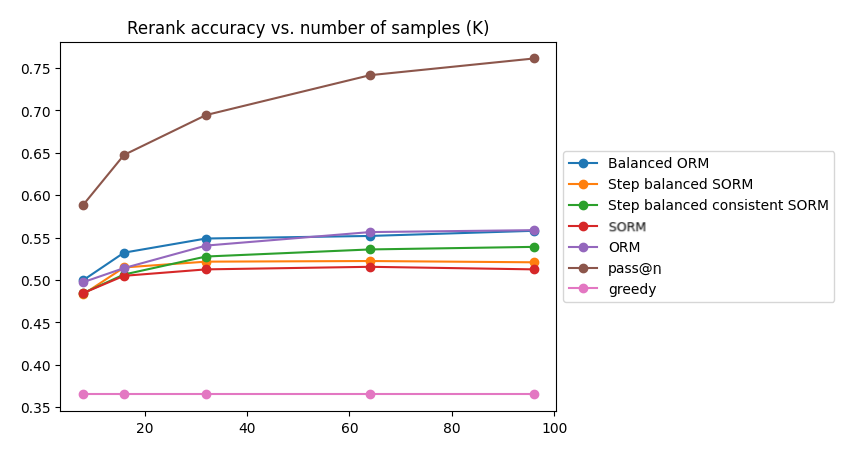}
    \caption{Plot of ORM, balanced ORM, and SORM rerank accuracies with the same SFT student (maj@1 = 0.36). Note: SORM by itself does not use balanced step labels or consistent verifiers as additional pre-processing steps as described in Section \ref{sec:method}. When we add in both steps, reranking performance significantly improves to nearly match the ORM's performance.}
\label{fig:weak_sft_rerank}
\end{wrapfigure}

Unfortunately, the inaccuracy of the SORM as a final answer predictor also makes it slightly worse as a final answer reranker. For this reason we always use the ORM whenever reranking candidate drafts and refinements. A more detailed comparison of reranking accuracies on GSM8K is done in Figure \ref{fig:weak_sft_rerank}. Note, this comparison is done using ORMs and SORMs derived from a student model trained using only supervised fine-tuning on GSM8K. Rerank accuracies are computed by sampling the student $K$ times and scoring each rollout with the ranker. The rollout with the highest score is then chosen as the final answer.

Figure \ref{fig:weak_sft_rerank} also plots rerank accuracies for SORM models trained on data without additional postproccessing. The best performing SORM uses only consistent verifying rollouts and per-step balanced labels, justifying these as good postprocessing choices.




\subsection{Evaluating global and local refinements}

\begin{figure}
    \centering  
    \includegraphics[scale=0.5]{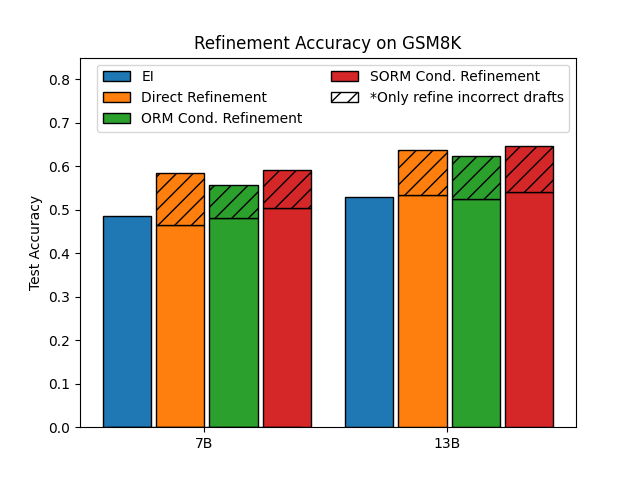}
    \includegraphics[scale=0.5]{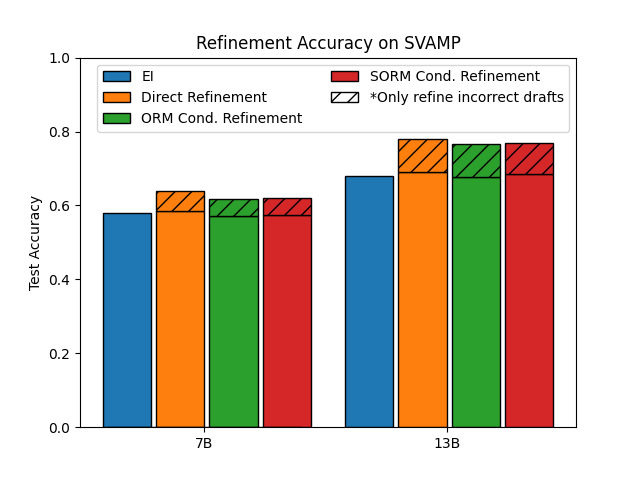}
    \caption{Refinement accuracies on GSM8K and SVAMP. All refinement models struggle identifying correct drafts which do not need refinement. Significant improvements are seen when only refining incorrect drafts.}
    \label{fig:single_refine}
\end{figure}

Now, with a better understanding of our SORMs' capabilities, we can apply them for refinement. Recall that to decide \textit{when} to accept a refinement $A_R$ we use the ORM as a reranker on the draft $A_D$ and refinement $A_R$. When performing local refinement we can additionally use both the ORM and SORM to identify the location of the first mistake in $A_D$. For the ORM we do this by labeling the first step $S_i$ such that $ORM(S_i) \leq T = 0.5$ where $T$ is a threshold hyperparameter. We identify the first error analogously with the SORM. We report results on both GSM8K and SVAMP $(Q, D)$ test sets in Figure \ref{fig:single_refine}. Note, we being evaluation without using the ORM as a reranker. This is done to confirm others' observations that refiners struggle knowing when to refine on their own.

\textbf{Both global and local refinement models struggle with knowing when to refine:} On both benchmarks global and local refinements show little improvement to overall model accuracy. GSM8K 7B global refinements even decreases overall accuracy, with the other models improving by at most 1\%. The local refinements improve overall accuracy more, likely due to the presence of the ``[BAD]" token indicating the location (and therefore presence) of the first mistake. This underscores the importance of an ORM for choosing when to refine an incorrect draft. We also note that bigger models produce better refinements.

\textbf{Global and local refinements fix similar percentages of incorrect drafts:} To understand how well our refiners perform when refinement is needed we also report results when applying refinement to only incorrect drafts from the test set in Figure \ref{fig:single_refine}. In this case both global and local refinements do much better, improving overall accuracy by an average of 10\% on GSM8K and 8\% on SVAMP. This demonstrates the refiners have learned how to refine, they simply often do not know when.

It is initially somewhat surprising global refinements are able to fix a similar percentage of drafts as local refinements. Local refinements receive extra information from $E$, presumably strictly improving performance over the global refiner. In reality, the provided $E$ is noisy as it must be predicted by an imperfect ORM/SORM. We see that even the difference in label accuracy bewteen the ORM and SORM results in a nontrivial difference in refinement accuracy. 

Additionally, global refinements have the advantage of optionally restarting a solution from scratch. A local refinement model is trained to reuse the prefix of a solution preceding a ``[BAD]'' token under the assumption this prefix has no errors. However, even if this prefix has valid reasoning, it may be a \textit{low-value} solution path for the student. For example, a student who often fails to correctly divide may benefit from starting the problem from scratch in a way that doesn't require any use of division. global refinements can take advantage of this, whereas local refinements may be commited to valid reasoning with a low chance of successfully completing. See Figure \ref{fig:refinements} for examples illustrating this point.

\textbf{Global and local refinements solve partially disjoint, complementary sets of problems:} To better understand how global and local refinements compare we examine the overlap between the problems they correctly solve. The last two rows of Table \ref{tab:incorrect_refine} show that, \textbf{when combined, global and local refinements can fix 41\% of incorrect GSM8K drafts} from the 13B student. Alone, global refinement and local refinement with the SORM fixes only 28\% of problems. Yet, when taking the best of both types of refinement for the same question, we significantly improve performance across all combinations of benchmarks and model sizes. This shows local refinement is able to solve a large set of problems global refinement cannot, and vice versa. Best performance at test time can then be achieved if we have a way of selecting which of the two refinements is appropriate.

\begin{table*}[t]
    \centering
    \begin{tabular}{@{}lrrrr@{}}
        & GSM8K 7B & GSM8K 13B & SVAMP 7B & SVAMP 13B \\
        \toprule
        Global Refinement & 0.203 & 0.281 & 0.14 & 0.255 \\
        Local Refinement + ORM & 0.182 & 0.262 & 0.09 & 0.229 \\
        Local Refinement + SORM & 0.211 & 0.283 & 0.11 & 0.237 \\
        Global Refinement + Local Refinement + ORM & 0.252 & 0.384 & 0.173 & 0.35 \\
        Global Refinement + Local Refinement + SORM & \textbf{0.280} & \textbf{0.412} & \textbf{0.19} & \textbf{0.37} \\
        \bottomrule
    \end{tabular}
    \caption{Refinement accuracy on incorrect model answers. Local refinement + SORM denotes using the SORM to highlight the first incorrect reasoning step for the local refinement model. We find refining both globally and locally with the SORM can fix up to 41\% of problems the model previously failed.}
    \label{tab:incorrect_refine}
\end{table*}

\begin{figure}
    \centering
    \includegraphics[scale=0.5]{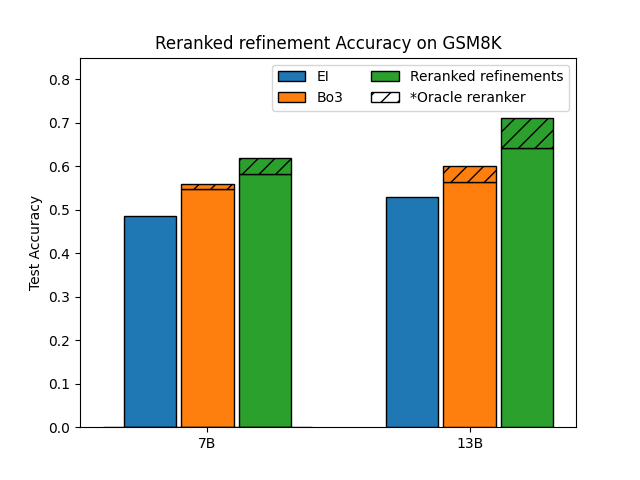}
    \includegraphics[scale=0.5]{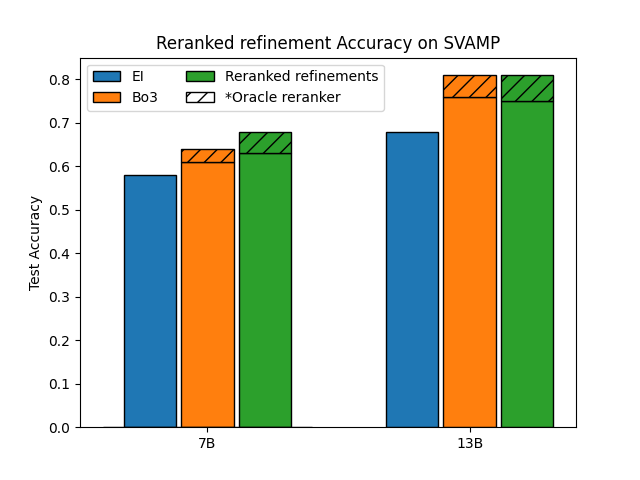}
    \caption{Accuracy of reranked refinements on all drafts compared to greedy and best of 3 samples from the student (Bo3) baselines. On GSM8K, reranking refinements using the ORM improves over the Bo3 baseline by up to 9\% and up to 13\% with a perfect reranker.}
    \label{fig:reranked_refine}
\end{figure}

Fortunately, we can use the ORM as a reranker for exactly the task of choosing between global and local refinements. Additionally, we can consider the initial draft as a third possible option as a way of deciding if we want to refine at all. Figure \ref{fig:reranked_refine} shows the results of reranking the draft, global, and local refinement for each question. Since we are effectively sampling three times, we include as a baseline the best of three (\textbf{Bo3}) samples from the EI student. We additionally report overall accuracy if we had a perfect reranker capable of always choosing the correct solution.

Reranking the draft + refinements improves over the draft accuracy by on average 8\% across models and benchmarks. When comparing with the Bo3 baseline we still see significant improvements of around 8\% on GSM8K. On SVAMP, reranked Bo3 is a much more competitive baseline, itself giving a large improvement over the draft accuracy. An even bigger improvement can be seen when using an oracle reranker, with the 13B refiner improving 11\% over even Bo3 on GSM8K.

\section{Conclusion and Future Work}

In this paper we study the use of reward models for both identifying \textit{when} to refine and \textit{where} to refine LLM reasoning. We found ORM models generalize to some extent to evaluating the accuracy of intermediate steps on easier reasoning tasks but struggle on harder tasks where the training data generating policy $\pi$ is further from $\pi^*$. We then propose to approximate the optimal policy $\pi^*$ via rejection sampling and post-processing, allowing us to generate training labels for intermediate steps $S_i$ used to train SORM models. We find the SORM generalizes better on intermediate test steps than the ORM, but at the cost of final answer accuracy. We then reused the ORM/SORM training data to train a global/local refinement models. We found each type of refinement strategy helped solve a largely unique set of problems, allowing us to combine both via ORM reranking for best performance. 

Future work can be classified as either: 1) improving the reliability and verbosity of local error critiques $E$ by providing more information on \textit{how} to refine or 2) augmenting the type of information local refiners use to generate correct solutions. Our study of both ORMs and SORMs reveals large room for improvement when verifying step level reasoning. Allowing verifier models to generate chains of thought appears to offer some benefit \citep{Dhuliawala2023ChainofVerificationRH}. Further augmenting verifying CoT with tools \citep{Zhou2023SolvingCM} allows GPT-4 to effectively solve MATH \citep{hendrycks2021measuring}. But it remains unclear how much GPT-4 relies on the tool to solve the problem versus actually uses the tool to augment its own understanding of \textit{why} a step is wrong.

Another promising direction treats iterative refinement as a form of \textit{in-context exploration} similar in spirit to ideas from algorithm distillation \citep{Laskin2022IncontextRL}. Here, the aim is to minimize the number of in-context model rollouts needed to figure out \textit{how} to refine. This also closely relates to work aiming to augment the exploration abilities of SOTA LLMs, a direction we believe is critical to future success. 
The right iterative local self-refinement strategies might hopefully allow models to access complex behaviors previously inaccessible with naieve iid repeated sampling.

\bibliographystyle{plainnat}
\bibliography{refs}

\newpage

\appendix


\section{Hyperparamters}
\label{sec:hparams}

\begin{table}[t]
    \centering
    \begin{tabular}{@{}lrrr@{}}
        & Expert Iteration & (S)ORM & Refiners \\
        \toprule
        Epochs & 4 & 1 & 1 \\
        max lr & 2e-5 & 2e-6 & 2e-5 \\
        min lr & 2e-7 & 2e-7 & 2e-7 \\
        Batch size & 128 & 256 & 128 \\
        \bottomrule
    \end{tabular}
    \caption{Hyperparameters for all training jobs. A cosine decay lr schedule is used in all cases.}
    \label{tab:hparams}
\end{table}

See Table \ref{tab:hparams} for a list of training hyperparameters used in each training job.

\section{RL for Reasoning}
\label{sec:rl_reasoning}

\begin{table*}[t]
    \centering
    \begin{tabular}{@{}lrrrr@{}}
         & maj@1 & maj@96 & Rerank@96 & pass@96 \\
        \toprule
         GSM8K & & & & \\
         \midrule
         SFT 7B & 0.41 & 0.47 & 0.54 & 0.72 \\
         SFT 13B & 0.48 & 0.55 & 0.68 & 0.84 \\
         EI$_2$ 7B & 0.485 & 0.55 & 0.64 & 0.8 \\
         EI$_2$ 13B & 0.53 & 0.59 & 0.71 & 0.88 \\
         \midrule
         SVAMP & & & & \\
         \midrule
         EI$_5$ 7B & 0.58 & 0.6 & 0.62 & 0.70 \\
         EI$_5$ 13B & 0.69 & 0.75 & 0.78 & 0.93 \\
         \bottomrule
    \end{tabular}
    \caption{Performance metrics for EI fine-tuned models on GSM8K and SVAMP. The subscript $n$ denotes the number of rounds of expert iteration until convergence of the maj@1 score. Reranking done via an ORM trained with samples from the model being reranked.}
    \label{tab:ei_results}
\end{table*}

In order to start from the best student possible we RL fine-tune the base-model using Expert Iteration. See Table \ref{tab:ei_results} for maj@1 (greedy), maj@96, Rerank@96 and pass@96 scores for EI fine-tuned models.

\section{RL for (global) refinement}
\label{sec:rl_global_refinement}



\textbf{Setup:} We compare the utility of PPO versus EI for refinement on the GSM8K benchmark. To train EI models we sample the SFT$^{2}$ model trained in Section \ref{sec:reasoning-rl} $K = 96$ times per prompt in the train set. We then pair together all incorrect solutions $A_{\textup{wrong}}$ with correct solutions $A_{\textup{correct}}$ for a fixed question $Q$ to form training tuples $(Q, A_{\textup{wrong}}, A_{\textup{correct}})$. We then fine-tune from Llama-2 7B to predict $p(A_{\textup{correct}} | Q, A_{\textup{wrong}})$ with the standard cross-entropy loss for a single epoch. We use an initial learning rate of 5e-5 decaying to 5e-7.

We initialize the PPO model from the SFT$^{2}$ checkpoint used in Section 2 above and use the same PPO parameters as when fine-tuning from the SFT checkpoint on GSM8K. A single example, included in the appendix for reference, is used to prompt the model for refinement. During training the student model is given a question $Q$ and draft $D$, where the draft is generated by the SFT model, and tasked with generating a refinement $R$. We give $R = \textbf{1}_{\texttt{is\_correct(R)}}-\textbf{1}_{\texttt{is\_correct(D)}}$ as a sparse reward at the end of the rollout. We additionally experimented with mixing in scores from an ORM, giving a final reward of $R = max(\textbf{1}_{\texttt{is\_correct(R)}}-\textbf{1}_{\texttt{is\_correct(D)}}, ORM(R) - ORM(D))$.

\subsubsection{Results for global refinement}

We evaluate all refinement models on a test set with questions from GSM8K test and drafts generated by SFT$^{2}$. Results are reported in Table \ref{tab:ppo_refine_acc}. Sampling at test time is done greedily, so only maj@1 accuracy is reported. We additionally report a 1-shot prompted Llama-2 7B as a baseline.

\begin{table}[t]
    \centering
    \begin{tabular}{@{}lrr@{}}
        & Accuracy \\
        \toprule
        SFT & 0.36 \\
        Prompted & 0.15 \\
        PPO & 0.36 \\
        ORM PPO & 0.36 \\
        EI & 0.39 \\
        \bottomrule
    \end{tabular}
    \caption{global refinement accuracies for prompted, PPO, and EI models. Note, maj@1 accuracy is reported for entire test set containing both correct and incorrect SFT generated drafts.}
    \label{tab:ppo_refine_acc}
\end{table}

\textbf{All models struggle learning when to refine} Our best refinement model is the EI model. However, EI improves over the SFT baseline by only 3\%, with PPO showing no improvement at all. This is because both models struggle correctly deciding when to refine. Often the EI model chooses to incorrectly refine correct drafts. The prompted pretrained model does even worse, having not been trained on GSM8K and thus struggling with when to refine and how to produce correct alternative refinements.

The PPO model collapses to simply returning the draft final answer, at least avoiding any negative rewards from incorrectly refining a correct draft. The prompted baseline also exhibits this copying behavior, accounting for the majority of its nontrivial accuracy. We experiment with alternative RL setups for preventing this degenerate behavior by using the ORM as an additional reward, removing the penalty for refining correct drafts, and/or having the student generate both the draft $D$ and refinement $R$. However, in all cases the model's copy bias continues to limit exploration, causing a collapse of the refinement to the initial draft.


\textbf{Discussion:} The results above highlight several failure modes. Firstly, models struggle with determining \textbf{when to refine}, often defaulting to never refining at all. Secondly, when the model does correctly choose where to refine it still struggles with knowing \textbf{where to refine}. Finally, even  knowing when and where to refine, the model still must decide \textbf{how} to refine. 

In order to improve model refinement capability we propose to decompose the problem by using unique models to solve each failure mode. Fortunately, deciding when to refine can naturally be handled by the ORM which is explicitly trained to predict when a final answer is correct. Additionally, when doing local refinement, we can use the SORM to identify where to refine. This now only requires the refinement model we train to decide how to refine, making the task significantly easier.

\section{Misc. Objectives for Reranking}

In \citet{Lightman2023LetsVS} the PRM is used for reranking by estimating $P(\texttt{Good}|S_i)$ for each step $S_i$ and taking the product. Inspired by this, we experimented with a number of different weightings for SORM intermediate step estimates when doing final answer reranking. For a solution of the form $S = (S_1, ..., S_L)$ these heuristics included:
\begin{enumerate}
    \item Final: $ORM(S_L)$
    \item Mean: $\frac{1}{L}\sum_{i=1}^L ORM(S_i)$
    \item Weighted mean: $\sum_{i=1}^L \frac{1}{L-i-1} ORM(S_i)$
    \item Min: $min_{i \in [L]} ORM(S_i)$
    \item Product: $\prod_{i=1}^L ORM(S_i)$
    \item Penultimate mean: $\frac{ORM(S_{L-1}) - ORM(S_L)}{2}$
\end{enumerate}

\begin{figure}
    \centering
    \includegraphics[scale=0.5]{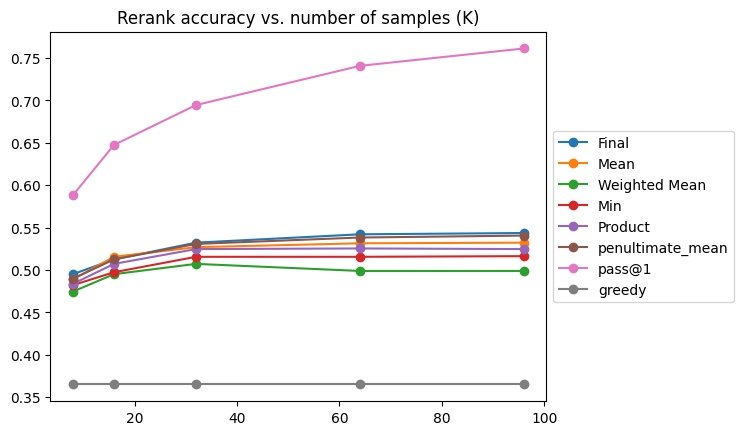}
    \caption{Comparison of heuristics for determining final rerank score.}
    \label{fig:combination_rerank}
\end{figure}

The results are plotted in Figure \ref{fig:combination_rerank}. Overall using only the final ORM estimates gives the best reranking accuracy with the penultimate mean coming in at a close second. The weighted mean significantly underperforms all other strategies, even taking the minimum ORM estimate.

\section{ORM and SORM extra-model generalization}



Both ORMs and SORMs exhibit signs of overfit to the data generating student $\pi$. When evaluated on GSM8K train set a 7B ORM model incorrectly classifies 42\% of correct solutions as incorrect. To examine this more closely we take two base student models, EI (trained with expert iteration) and SFT (supervised fine-tuning), and use both to generate training data for $ORM_\textup{EI}$ and $ORM_\textup{SFT}$ respectively. we then evaluate both ORMs on test sets generated by each model. Results are reported in Table \ref{tab:ei_sft_orm}. We find both ORMs underperform on the test dataset generated by the opposite student model.

\begin{table}[t]
    \centering
    \begin{tabular}{@{}lrr@{}}
        & EI Test Accuracy & SFT Test Accuracy \\
        \toprule
        $ORM_\textup{EI}$ & 0.64 & 0.51 \\
        $ORM_\textup{SFT}$ & 0.58 & 0.56 \\
        \bottomrule
    \end{tabular}
    \caption{Evaluation $ORM_\textup{EI}$ and $ORM_\textup{SFT}$ cross-generalization.}
    \label{tab:ei_sft_orm}
\end{table}

\section{Contrastive vs. Classifier RMs}

Both the ORM and SORM are trained as classifiers to predict the probability of a $\texttt{good}$ label at each intermediate step $S_i$. However, in RLHF there are only preference comparisons over solutions. So the RLHF reward model is often trained via a contrastive loss $-log(\sigma(RM(y_\texttt{good}) - RM(y_\texttt{bad}))$. We explore the use of a contrastive reward model in the reasoning setting, comparing reranking performance with the ORM. Training data is sampled from a 7B SFT student model with K = 96 rollouts per training prompt at a temperature T = 1.0. We assign ORM step labels in the usual way, setting $l_i = 1$ at a step $S_i$ if $l_L = 1$ and otherwise $l_i = 0$. To construct preference pairs we select the maximal equal number of positive and negative solutions for the same prompt and form pairs $(S_\texttt{good}, S_\texttt{bad})$ with no solution repeated. Only the contrastive loss on the final token is backpropagated. We then rerank solutions on the test set using scores assigned by both the classifier ORM and contrastive ORM. We find the classifier ORM gets $0.56$ rerank accuracy whereas the contrastive ORM gets $0.47$, \textbf{suggesting the classifier to be a strictly better reranker}.

\section{Accuracy of the SORM data generating method}
\label{sec:sorm_data_acc}

The SORM data generation process will suffer from both false positives and false negatives. False positives may occur when the student model solves the problem incorrectly but gets the right final answer. False negatives will occur when the rejection sampled student is simply unable to solve the problem from a prefix $P_i$ desipte the prefix being logically valid. In order to verify the correctness of the SORM step-level verification process we hand-label several model generated solutions on GSM8K and compute how well the our ground truth labels align with the generated labels. Over $n = 64$ and a total of $257$ steps we find the SORM data labels agree with our ground truth 94\% of the time.

\section{Mixing other sources of PRM data}

We additionally experiment with mixing in PRM data on the MATH dataset from \citet{Lightman2023LetsVS}. We train Llama-2 7B to predict $\texttt{negative}$,  $\texttt{neutral}$ and $\texttt{good}$ labels for each step, with ``bad'' steps being incorrect, ``neutral'' steps neither making forward nor backward progress, and ``good'' steps being correct and useful to solving the problem. The resulting PRM gets $0.91$ accuracy on the MATH PRM test set. However, we find the PRM transfers poorly as a final answer correctness predictor, getting only $0.58$ accuracy on an EI generated test set.

\section{Self-supervised learning with the SORM}

It is likely the SORM dataset generation process is fairly noisy. Low quality samples will directly impact the performance of the downstream SORM, making it critical to improve dataset quality. In an attempt to remove noisy outliers we filtered version of the SORM dataset via SORM self-supervsion. For each training pair $(Q, P_i, l_i)$, where $Q$ is the question, $P_i = (S_1, ..., S_i)$ is a solution prefix with $i$ steps, and $l_i$ is the correctness label, we apply $SORM((Q, P_i))$. This generates a self-supervised label $l_i' = \textbf{1}_{SORM((Q, P_i)) > 0.5}$. We then filter out all training samples with $l_i' \neq l_i$. 

We filter the SORM dataset with a SORM checkpoint trained for 1 epoch and another trained for 2 epochs. The first model, denoted as SORM$_1$, has 75\% accuracy on the SORM test set but 91\% on the SORM train set. SORM$_2$ gets 78\% test but 95\% on the train set. It could be SORM$_2$ is overfit to the train set, so we train new SORM models on both filtered datasets. SORM$_1'$, trained on SORM data filtered with SORM$_1$, gets 79\% accuracy. SORM$_2'$, trained on SORM data filtered with SORM$_2$, gets the same. 


\section{Value refinement}
\label{sec:value_refinement}

The local refinement strategy employed in the main body of this work uses critiques attempting to locate steps with logical errors. This can be interpreted as a type of \textit{process-based refinement} which which gives feedback agnostic to the abilities of the refinement model. More sophisticated forms of feedback might take the underlying capabilities of the model into account, maximizing the chances of this particular student succeeding.

One alternative refinement strategy which gives student specific feedback is \textit{value-based refinement}. A value-based refinement model receives feedback in the form of a ``[BAD]'' token at the step in a solution with the highest value for the model. Recall the value of a step $S_i$ is the probability the model gets the correct answer from $S_i$. Note this is not the same thing as process-based feedback as the step $S_{i+1}$ after the highest value step $S_i$ may not necessarily contain an error. Instead, $S_{i+1}$ may attempt to solve the problem is a difficult way, for example using division with a model which struggles dividing correctly.

\textbf{Training a value-based refinement model} Further recall the ORM directly estimates the value function $V^\pi$ of its data generating policy $\pi$ such that $ORM(P_i) \approx V^\pi(S_i)$ for a prefix with $P_i = (S_1, ..., S_i)$. Given a student model $\pi$ on a reasoning task $\tau$ we generate an ORM training set $\mathcal{D}_{ORM}$ by sampling each prompt in $\tautrain$ $K = 96$ times. We train the ORM as a classifier, setting an intermediate step label $l_i = l_L$ where $l_L = \texttt{is\_correct(S)}$. 

To construct the value based refinement dataset $\mathcal{D}_{\textup{vrefine}}$ we start by reusing the SORM dataset $\mathcal{D}_{\textup{SORM}}$ generated as above using policy $\pi$ and rejection sampling. For a sample $S = (S_1, ..., S_L)$ we identify the highest value step $S_i$ by choosing the step with the most correct verifying rollouts $v_i^j$. We then select one of the verifying rollouts whose first step differs from $S_{i+1}$ as the improving refinement $R$. This forms a value-refinement training pair $(Q, S, R, C)$ where $C$ is a ``[BAD]'' token inserted before step $S_{i+1}$ in $S$. We then train a value-based local refinement model by minimizing $p(R | Q, S, C)$ with the standard cross-entropy loss.

In practice we generate all downstream models and datasets using LLama-2 7B EI on GSM8K as $\pi$.

\textbf{Results:} To evaluate we use the 7B EI model trained on GSM8K from our best SFT checkpoint. We greedily sample solution drafts on the GSM8K test set, forming a $(Q, D)$ test set for the value-based local refinement model. We then label the highest value step $S_i$ of each draft using the ORM, placing the ``[BAD]'' token as a prefix to $S_{i+1}$. 

We evaluate only on incorrect drafts, comparing directly the performance of process-based SORM refinement. Value-based refinement fixes 14\% of incorrect drafts whereas the SORM baseline fixes 21\%. Surprisingly, even global refinement outperforms value-based refinement by 6\%. This again take this to suggest intermediate ORM estimates are fairly noisy on test data.


\end{document}